\documentclass[letterpaper, 10 pt, conference]{ieeeconf}  

\IEEEoverridecommandlockouts                              

\usepackage{graphics} 
\usepackage{epsfig} 
\usepackage{amsmath} 
\usepackage{amssymb}  
\usepackage{bm}  
\usepackage{booktabs}
\usepackage{graphicx}
\usepackage[caption=false]{subfig}
\usepackage{color}
\usepackage{enumerate}
\usepackage[linesnumbered,ruled]{algorithm2e}
\usepackage{multicol}
\usepackage{siunitx}
\usepackage{comment}
\usepackage{dblfloatfix}

\title{\LARGE \bf
Unified Multi-Contact Fall Mitigation Planning for Humanoids via Contact Transition Tree Optimization}

\author{Shihao Wang$^{1}$ and Kris Hauser$^{2}$
\thanks{*This work was supported by NSF grant NRI \#1527826}
\thanks{$^{1}$Shihao Wang is with the Department of Mechanical Engineering and Materials Science, Duke University,
        Durham, NC 27708, USA
        {\tt\small shihao.wang@duke.edu}}%
\thanks{$^{2}$Kris Hauser is with the Departments of Electrical and Computer Engineering and Mechanical Engineering and Materials Science, Duke University, 
        Durham, NC 27708, USA
        {\tt\small kris.hauser@duke.edu}}%
}

\begin{document}
\setlength{\abovedisplayskip}{3pt}
\setlength{\belowdisplayskip}{3pt}
\maketitle
\thispagestyle{empty}
\pagestyle{empty}

\begin{abstract}
This paper presents a multi-contact approach to generalized humanoid fall mitigation  planning that unifies inertial shaping, protective stepping, and hand contact strategies.  The planner optimizes both the contact sequence and the robot state trajectories. A high-level tree search is conducted to  iteratively grow a contact transition tree.  At each edge of the tree, trajectory optimization is used to calculate robot stabilization trajectories that produce the desired contact transition while minimizing kinetic energy. Also, at each node of the tree, the optimizer attempts to find a self-motion (inertial shaping movement) to eliminate kinetic energy. This paper also presents an efficient and effective method to generate initial seeds to facilitate trajectory optimization. 
Experiments demonstrate show that our proposed algorithm can generate complex stabilization strategies for a simulated planar robot under varying initial pushes and environment shapes.
\end{abstract}

\section{INTRODUCTION}

Humanoid robots have the apparent advantage of being able to navigate using legs and arms to traverse terrains which are demanding for wheeled robots. However, the intrinsic instability of bipedal walking makes them much harder to control than wheeled robots, and falls can cause costly failures. As a result, fall mitigation is a topic of active research.  Prior approaches have either used the internal joints of the robot to resist disturbances (e.g. ankle strategy, hip strategy \cite{Stephens2007,Aftab2012}, or inertial shaping \cite{Goswami2014}) or external contacts (e.g., protective stepping \cite{Pratt2006}, knee contact \cite{Fujiwara2006}, hand contact \cite{Humanoid2017}, \cite{Wang2018}).   However, these strategies have been considered in isolation and there have been limited attempts to unify multiple strategies into a single fall mitigation system. Existing unified strategies use heuristic decision functions for mode switching \cite{Ogata2008,Kamioka2017}, and machine learning-based methods for fall planning \cite{SehoonHa2015}, \cite{Yi2013}. Past approaches can only handle a limited range of conditions, due to oversimplified assumptions of robot dynamics and contact, or a restrictive number of protective strategies.

In this paper we consider the following question: what is the {\em optimal} strategy by which a robot should mitigate an impending fall?  Small pushes can be easily resisted by inertial shaping, while larger pushes may require one or more protective steps, and possibly even hand contact.  Our hypothesis is that robot should choose the {\em strategy that yields the fastest decrease in kinetic energy}. 

\begin{figure}[!t]
    \centering
    \includegraphics[scale = 0.36]{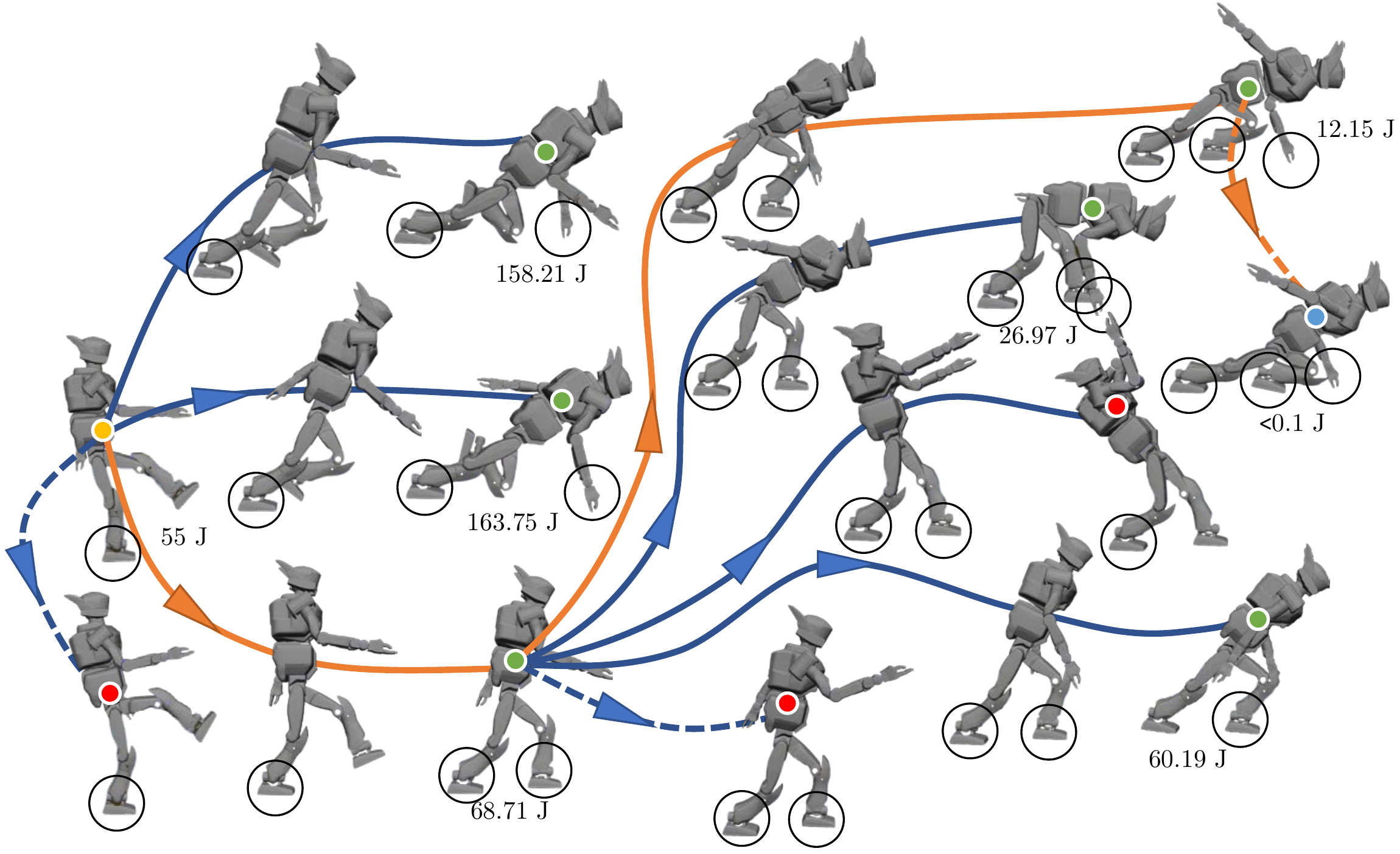}
    \caption{\small A representative contact transition tree for fall mitigation with initial kinetic energy \SI{55}{\J} where the robot stabilizes itself with protective stepping, hand contact and inertial shaping. Dots indicate root node (yellow), terminal node (blue),  connectable nodes (green), and unconnectable or unstabilized nodes (red). Dashed curves denote self-motion (inertial shaping) trajectories. Note that intermediate configurations of transition motions are also shown.  Orange curves indicates the solution, with contact sequence RF $\to$ RF/LF $\to$ RF/LF/LH.}
    \label{fig:Flat_KE55}
\end{figure}

In order to study this hypothesis, we develop a planning method that simultaneously generates the contact sequence and optimized whole-body trajectories to achieve a stabilizing multi-contact trajectory.  With the assumption of a planar robot model in Sagittal plane, the proposed method obeys dynamics, collision impact mapping, and contact feasibility constraints, and can generate a diverse number of fall mitigation strategies, including flat ground and vertical wall, and allows the adoption of hand contact for fall stabilization.  Our algorithm iteratively constructs a contact transition tree, rooted at the robot's disturbed initial state, where each edge denotes a contact transition connected by a feasible dynamic trajectory.  It proceeds by attempting to optimize a self-stabilizing motion at the root node, without changing contact. If self-stabilization succeeds, then the algorithm is done.  If this fails, our algorithm expands to its adjacent contact modes and then attempts to optimize robot transition trajectories to reach these contact modes. Search proceeds via a minimum kinetic energy heuristic, and continues until a self-stabilizing motion can be found at some node.  The path from the root to this node yields a multi-contact trajectory that stabilizes the robot. Fig.~\ref{fig:Flat_KE55} illustrates one representative diagram of a contact transition tree.

For trajectory optimization we use the direct collocation method to simultaneously optimize robot configuration, velocity, control and contact force \cite{Posa2016}.  We propose a seed initialization algorithm that calculates an initial guess for the optimizer. This strategy initializes the intermediate robot states with an evenly evaluation of the computed parabolic splines matching initial robot state and an optimized feasible goal configuration, and computes the control and contact forces with a pseudo-inverse method to minimize the dynamics constraint violation. Experiments with a simulated planar robot demonstrate that our proposed algorithm automatically generates unified mitigation strategies, such as inertial shaping, protective stepping and hand contact, to stabilize varying initial pushes and environment shapes.

\section{Related Work}
The problem of balancing a biped or humanoid in response to external disturbance has been an active topic of research for some time~\cite{Fujiwara2006,Pratt2006}. Strategies proposed to address this topic can be classified into two main categories: {\em fixed contact} and {\em contact modification}. Fixed contact strategies aim to recover the robot purely through joint effort to regulate linear and angular momentum, all while maintaining the current contact state. Two representatives are the ankle strategy and hip strategy \cite{Stephens2007}\cite{Aftab2012}.  For larger disturbances it is impossible to recover to a stationary state using joint torques alone. Contact modification strategies reduce momentum by making contact at the robot extremities, which transfers kinetic energy from the robot into the environment. Examples of this type of strategy include protective stepping \cite{Stephens2010}, hand contact \cite{KunihiroOgata2007} \cite{Marcucci2017}, knee contact \cite{JiuguangWang2012}, tripod posture \cite{Yun2014}, and contact with accessories such as a backpack \cite{Lee2011} and walking sticks \cite{Tam2016}. 

Contact modification strategies may also be divided into push recovery and fall mitigation approaches, which differ in whether the robot recovers to a normal operational state after the disturbance. \emph{Push recovery} assumes a moderate external disturbance and the robot is assumed capable of regulating its locomotion to decrease the increased momentum. Capture point stepping is a commonly used strategy which aims to dampen the robot centroidal velocity by making additional footstep(s) \cite{Koolen2012Capturability}. This strategy employs an inverted pendulum model with a massless telescopic leg and produces analytic information of the capture steps. This strategy has been extended on omnidirectional disturbance \cite{Missura2013}, uneven terrain stepping \cite{Ramos2015} and model validation analysis \cite{PosaKT17}. \emph{Fall mitigation} deals with large disturbances such that an unavoidable fall has been triggered, and plans the contact and fall trajectories of the robot to reduce the robot damage from collision impulse. Strategies focusing on the impact reduction in pre-impact stage adopt the damage optimization with simplified model using indirect Pontryagin’s minimum principle \cite{Fujiwara2007} and direct collocation method \cite{JiuguangWang2012}, multi-contact planning for whole-body trajectory \cite{SehoonHa2015} and posture reshaping to avoid configuration singularity \cite{Luo2016}. Other strategies focus on the post-impact stabilization with active compliance adjustment \cite{Hoffman2013} \cite{Samy2017} and the optimization of both pre/post-impact stage to reduce damage from impulse and potential contact slippage \cite{Humanoid2017} \cite{Wang2018}.  Planning fall mitigation motions is generally more computationally challenging due to greater diversity of initial conditions, a wider range of possible contact sequences, and the difficulty of devising simplified dynamic models that are suitable to use when the robot is far from nominal operating conditions.


Despite the existence of various disturbance recovery strategies, it still remains unclear which strategy or combination of strategies should be adopted to stabilize a humanoid if an arbitrary pushed is imposed. Stephens determines the decision boundary between strategies using a simplified LIPM model \cite{Stephens2007}. Our proposed method unifies both fixed contact and contact modification strategies, and can also devise novel contact sequences. It does so by planning trajectories from the initial state to minimize kinetic energy via a multi-contact transition tree approach. 
This approach is related to other multi-contact planning algorithms, such as manipulation planning in contact configuration space~\cite{Ji2001}, multi-modal motion planning for legged robots \cite{Hauser2010}, and robot whole-body transition synthesis using motion capture dataset \cite{Mandery2015b}. However, these approaches only address quasi-static systems and feasible planning.  Our approach is based on a planar dynamic model of the robot and uses trajectory optimization to generate state-space paths.

\section{Method}

\subsection{Contact Transition Tree}
Given an initial robot state, initial contact mode, and environment geometry, we wish to generate a joint space trajectory and contact sequence to stabilize to a stationary (zero-velocity) state.  Our method integrates a high-level tree search, to explore contact sequences, with trajectory optimization, to plan connecting trajectories and self-stabilization trajectories. 

Let $\bm{x} = (q,\dot{q})$ denote a robot state and $\bm{\sigma}$ denote a contact mode, with $\bm{x}_0$ and $\bm{\sigma}_0$ the start state and mode.  A state consists of the robot's generalized configuration $q\in \mathbb{R}^{n}$ and velocity $\dot{q} \in \mathbb{R}^{n}$. A contact mode indicates the contact active/inactive status, and is a vector $\bm{\sigma} \in \{0,1\}^{l\times 1}$ where $l$ is the number of contact extremities allowable on the robot. $\bm{\sigma}_i=0$ indicates that the $i$'th extremity has no contact, while $\bm{\sigma}_i=1$ indicates contact.

A feasible {\em fixed-contact trajectory} must respect contact constraints of its mode, as well as dynamic constraints, friction limits, torque limits, and joint limits.  These constraints will be described in Sec.~\ref{sec:DynamicConstraints}. Let the kinetic energy of a state be $E_k(\bm{x}) = \frac{1}{2}\dot{q}^T D(q) \dot{q}$. We define a stationary state as one in which $E_k(\bm{x})$ is sufficiently small.  Our goal is to produce a {\em multi-contact trajectory} sequence of modes $\bm{\sigma}_0,\ldots,\bm{\sigma}_N$ and a continuous sequence of $N+1$ feasible trajectories starting at $\bm{x}_0$ and ending at $\bm{x}_N$.  $N$ is not fixed, and there is no restriction on the terminal contact mode. 

For transitions $\bm{\sigma}_{i} \rightarrow \bm{\sigma}_{i+1}$ that add a new contact, there is additionally an {\em impact mapping} condition that must be met at the transition to account for the instantaneous change in velocity. In this contact addition case, the pre-impact kinetic energy is the cost function to be minimized.


To build a feasible multi-contact trajectory, our method incrementally builds a {\em contact transition tree} $\mathcal{T}$, rooted at the initial robot state's mode, and iteratively grows its edges to the most promising stabilizable nodes until a terminal self-stabilization has been achieved. 
Each tree node $\mathbf{d}_i$ contains three attributes:
\begin{itemize}
    \item Contact mode $\bm{\sigma}(\mathbf{d}_i)$
    \item Robot state $\bm{x}(\mathbf{d}_i)$.
    \item Self-motion trajectory $y_{\text{self}}(\mathbf{d}_i)$.
\end{itemize}
The {\em self-motion trajectory} $y_{\text{self}}(\mathbf{d}_i)$, also denoted by $y_i$, is a feasible fixed-contact trajectory at $\bm{\sigma}(\mathbf{d}_i)$ starting at $\bm{x}(\mathbf{d}_i)$.  In other words, it is an inertial shaping trajectory.

\begin{algorithm}[!b]
\DontPrintSemicolon
\SetKwInOut{Input}{Input}
\SetKwInOut{Output}{Output}
\SetKwRepeat{Do}{do}{while}
    \Input{Initial state $\bm{x}_0$, mode $\bm{\sigma}_0$, environment map}
    \Output{Mode sequence: $(\bm{\sigma}_0 \to \bm{\sigma}_{1} \to \cdots \to \bm{\sigma}_N )$ \\
    Trajectories: $(y_{0,1} \to\cdots \to y_{N-1,N} \to y_{N})$ }
    $\mathbf{d}_0 \gets $\texttt{Node}$(\bm{x}_0,\bm{\sigma}_0)$ \;
    $\mathcal{T}\gets \mathbf{d}_0$, $\mathcal{F}\gets \{\mathbf{d}_0$\} \;
    \While{$|\mathcal{F}| > 0$}
    {
        $\mathbf{d}_i \gets$ \texttt{Pop}($\mathcal{F}$)\;
        $y_i\gets$ \texttt{opt\_self\_motion}$(\mathbf{d}_i)$\;
        \If{\upshape $y_i \neq nil$ and $E_k(\text{End}(y_i))<\epsilon_{E_{k}}$}
        {
            $y_{\text{self}}(\mathbf{d}_i) \gets y_i$\;
            Retrieve path from $\bm{d}_0$ to $\bm{d}_i$ in $\mathcal{T}$\;
            \Return Modes $(\bm{\sigma}_0 \to \cdots \to \bm{\sigma}_i)$\;
            \hspace{10mm} Paths $(y_{0,1} \to \cdots \to y_{i-1,i} \to y_i)$\;
        }
        
        \For{\upshape $\bm{\sigma}_{j} \in $ AdjacentModes$(\bm{\sigma}(\mathbf{d}_i))$}
        {
            {$y_{i,j}$ = \texttt{opt\_transition\_motion}($\mathbf{d}_i$, $\bm{\sigma}_{j}$)\;
            \If{$y_{i,j} \neq nil$}
            {
                $\bm{x}_j \gets \text{End}(y_{i,j})$ \;
                If $|\bm{\sigma}_j| > |\bm{\sigma}_i|$, $\bm{x}_j \gets \text{ImpactMap}(\bm{x}_j,\bm{\sigma}_j)$ \;
                $\mathbf{d}_j \gets \text{Node}(\bm{x}_j,\bm{\sigma}_j)$ \;
                Add $\mathbf{d}_j$ to $\mathcal{T}$ as a child of $\mathbf{d}_i$\;
                Store $y_{i,j}$ with $\mathbf{e}_{i\to j}$ \;
                Add $\mathbf{d}_i$ to $\mathcal{F}$ with priority $E_k(\text{End}(y_{i,j}))$\;
            }
            }
        }
    }
    \Return no solution found
    \caption{Contact transition tree search}
    \label{alg:1}
\end{algorithm}

For each edge $\mathbf{e}_{i \to j}$ from $\bm{d}_i$ to $\bm{d}_j$, the nodes may differ by exactly one limb in contact. Each edge also stores a {\em transition trajectory} $y_{ij}$ starting at $\bm{x}(\mathbf{d}_i)$ and terminating in $\bm{x}(\mathbf{d}_j)$.  A transition trajectory must satisfy the constraints of $\bm{\sigma}_i$. If the mode switch $\bm{\sigma}_i \rightarrow \bm{\sigma}_j$ removes a contact, then the final state must satisfy the dynamic constraints of $\bm{\sigma}_j$, specifically, that there are valid forces at the contacts active in $\bm{\sigma}_j$.  If the mode switch adds a contact, then the final state must satisfy the kinematic contact conditions of $\bm{\sigma}_j$

\subsection{Contact Transition Tree Search and Expansion}

The following search procedure is used to grow $\mathcal{T}$.  Let  $\mathcal{F}$ denote the {\em frontier} nodes, which is implemented as a priority queue sorted by increasing kinetic energy. 
\begin{enumerate}
    \item The node with lowest kinetic energy is extracted from $\mathcal{F}$, and a trajectory optimization will be conducted to calculate $y_i$.  
    \item If a stationary endpoint is found, then we are done.
    \item Otherwise, the algorithm continues to expand to neighboring contact modes by attempting to find feasible trajectories to those modes. 
    \item Each successful connection to a neighbor is added as a new edge in $\mathcal{T}$, and each neighboring node is added to $\mathcal{F}$.
\end{enumerate}
   
Specifically, our algorithm uses a greedy approach in which each trajectory optimization attempts to {\em minimize the kinetic energy of the endpoint}.  This approach tries to eliminate kinetic energy from the initial robot state as quickly as possible, which limits the amount of search needed to find a solution. 

Algorithm \ref{alg:1} illustrates the details of tree search and expansion procedure.  Lines 5--11 attempt to stop the robot at the current contact mode via a self motion.  If the optimization succeeds and the kinetic energy at the end point is within a small tolerance $\epsilon_{E_{k}}$, we are done.  The result traces back from this leaf node to the root node to extract the solution contact sequence and single-contact trajectory sequence.
Lines 12-22 validate the connectivity to child nodes using optimization.  If no further feasible paths can be found (Line 24), the algorithm terminates with failure.

The main computational tasks are undertaken by two subroutines:
\begin{itemize}
    \item \texttt{opt\_self\_motion} optimizes a robot stabilization trajectory at the contact mode $\bm{\sigma}$ of node $\mathbf{d}_i$.
    \item \texttt{opt\_transition\_motion} optimizes a transition trajectory from node $\mathbf{d}_i$ to node $\mathbf{d}_{\text{child}}$.
\end{itemize}
These will be described in more detail below.
\begin{figure}[!b]
    \centering
    \includegraphics[scale = 0.3]{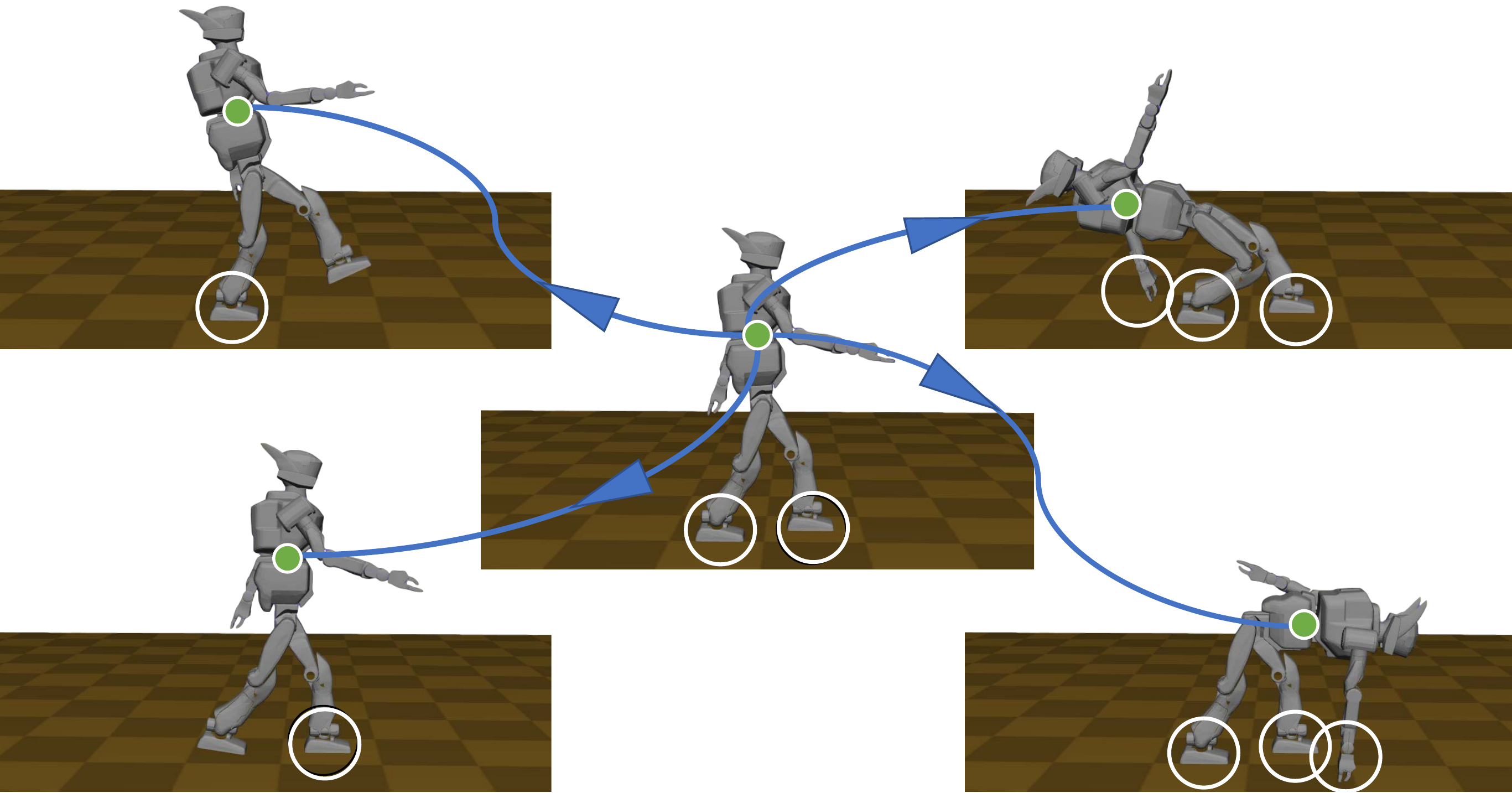}
    \caption{Illustrating the mode adjacency diagram. From two-foot contact mode LF/RF (center), the robot can switch to one-foot modes LF and RF (left) and two three-contact modes LF/RF/LH and LF/RF/RH (right).}
    \label{fig:node_expansion}
\end{figure}
The following minor subroutines are also used: 
\begin{itemize}
    \item Pop takes out the node with the minimum kinetic energy among other nodes in the Frontier $\mathcal{F}$.
    \item Node creates a node at a given robot state and contact mode.
    \item End returns the robot state at the end of a trajectory. 
    \item AdjacentModes produces the list of neighbouring contact modes that differ from $\bm{\sigma}$ by exactly one change of contact. Fig.~\ref{fig:node_expansion} illustrates a representative node expansion example where each hand/foot contact can be modified to produce 4 adjacent nodes.
    \item ImpactMap calculates the post-impact robot state resulting from impact mapping as described in Sec.~\ref{sec:ImpactMapping}.
\end{itemize}

\subsection{Trajectory Optimization: Stabilization and Transition}
Trajectory optimization is central to our method, and we use a collocation method that uses a high-accuracy spline representation~\cite{Posa2016}. As an objective function we minimize the kinetic energy at the end state of the trajectory. We also develop a custom trajectory initialization that works well in practice.  Both \texttt{opt\_self\_motion} and \texttt{opt\_transition\_motion}, use the same underlying method with only small modifications.

\subsubsection{Constraints}
\label{sec:DynamicConstraints}

The standard equation of motion for constrained dynamical system is
\begin{equation}
    D(q)\ddot{q} + C(q,\dot{q}) + G(q) = J(q)^{T}\bm{\lambda} + B\bm{u}
    \label{eqn:dynamics}
\end{equation}
where $D(q) \in \mathbb{R}^{n \times n}$ is the inertia matrix. $C(q,\dot{q}) \in \mathbb{R}^{n \times 1}$ is the centrifugal and coriolis matrix. $G(q) \in \mathbb{R}^{n\times 1}$ is the generalized gravitional matrix.  $\bm{u} \in \mathbb{R}^{m \times 1}$ is the joint torque vector and $B \in \mathbb{R}^{n \times m}$ is the input matrix. $\bm{\lambda} \in \mathbb{R}^{l \times 1}$ be the contact force vector.   $J(q) \in \mathbb{R}^{l \times n}$ is the Jacobian matrix of global contact positions with respect to $q$. 

This constraint must be met for all states along the trajectory. We also require the following feasibility constraints.
\begin{itemize}
    \item {\em Complementarity constraints}: The elementwise product between  force $\lambda_j$ at the $j$'th contact and the associated relative distance  $\bm{\phi}_j(q)$ has to be 0:
    \begin{equation}
        \lambda_{j} \cdot \phi(q)_j = 0, \forall j \in \left[1,...,l \right].
    \end{equation}
    But since the contact mode is known during optimization, we enforce stricter constraints on contact force $\bm{\lambda}_i$ and relative distance. Let $Diag(\cdot)$ generate a diagonal matrix with the vector $\cdot$ on its diagonal and let $\neg$ denote logical negation. Then the constraint is rewritten as 
    \begin{equation}
    \begin{split}
     Diag(\neg \bm{\sigma})&\bm{\lambda} = 0,\\
     Diag(\bm{\sigma})\bm{\phi}(q) = 0, \hspace{1mm} &Diag(\neg \bm{\sigma})\bm{\phi}(q) >\epsilon_{ct}.
    \end{split}
    \end{equation}
    The parameter $\epsilon_{ct}$ enforces a minimum clearance for non-contacting limbs. Note that $\epsilon_{ct}$ needs to be set to 0 if the transition is to remove a certain contact.
    \item {\em Contact holonomic constraints}: 
    In addition to contact position constraints, we also enforce contact velocities to be 0 using:
    \begin{equation}
        Diag(\bm{\sigma}) J(q)\dot{q} = 0.
    \end{equation}
    \item {\em Contact force feasibility constraints}: 
     The contact force between the robot and the environment follows the Coulomb friction model, and the friction cone constraints are written as
    \begin{equation}
    \lambda_{j\bm{n}} \geq 0, \hspace{1.5mm} \mu^2\lambda_{j\bm{n}}^{2} \geq \lambda_{j\bm{t}}^{2}
    \end{equation}
    where $\lambda_{j\bm{n}}$ and $ \lambda_{j\bm{t}}$ respectively denote the contact force in the normal and tangential direction of contact $j$.
    
    \item {\em Joint limits, velocity limits, and torque limits}:
    \begin{equation}
    \begin{split}
        &\bm{x}_{min} \leq \bm{x} \leq \bm{x}_{max}\\
        &\bm{u}_{min} \leq \bm{u} \leq \bm{u}_{max}
    \end{split}
    \end{equation}
\end{itemize}

\subsubsection{Direct collocation}
The variables to be optimized are the time duration $T$ and trajectories of the robot state, control and contact force. After the transcription of these continuous trajectories at $N_d$ equally distributed knot points with timestep $h = \frac{T}{N_d - 1}$, we formulate this trajectory optimization into a non-linear programming (NLP) problem. The inputs to the NLP are timestep $h$, discretized robot state $(\bm{x}_1, ..., \bm{x}_{N_d})$, control ($\bm{u}_1, ...,\bm{u}_{N_d}$) and contact force $(\bm{\lambda}_1, ..., \bm{\lambda}_{N_d})$. 

Due to second-order nature of the robot dynamical system and the holonomic constraints on contact position and velocity, using classic first-order Euler integration to update the robot state tends to cause numerical difficulties. This drawback can be avoided by approximating the robot state and control/contact force trajectories as implicit cubic splines and piecewise linear functions, respectively. A third-order integration accuracy $\mathcal{O}(h^3)$ has been reported with this spline choice \cite{Posa2016}. 
The construction of implicit cubic splines is associated with the system kinematics and dynamics. For a representative position variable $q_i$, its cubic spline path within a timestep can be expressed
\begin{equation}
    q_i(s) = a_ps^3 + b_ps^2 + c_p s + d_p, s \in [0 , 1]
\end{equation}
The position $q_i(s)$ and its first time derivative $\frac{dq_i(s)}{dt}$ should match the robot state at both edges $(\bm{x}_i, \bm{x}_{i+1})$. These matching conditions solve the four unknowns in $q_i(s)$ and any intermediate point can be then interpolated. However, the same methodology cannot be used to calculate the cubic spline coefficients of the velocity variable $\dot{q}_i(s)$ since the first order derivative of $\dot{q}_i(s)$, acceleration, is not a variable to be optimized. As a result, we have to adopt a different approach to get its cubic spline.

At sequential knots, states $(\bm{x}_i, \bm{x}_{i+1})$, controls $(\bm{u}_i, \bm{u}_{i+1})$ and contact forces $(\bm{\lambda}_i, \bm{\lambda}_{i+1})$ are optimization variables. Instead of enforcing the dynamics constraint (\ref{eqn:dynamics}) inside the optimization solver, we directly make use of this constraint to calculate the acceleration determined under the current set of robot state, control and contact force. With accelerations now available at both knots, the cubic spline coefficients of $\dot{q}_i(s)$ can be computed.

The guaranteed satisfaction of the dynamics constraints at knot points enable us to add a collocation point in the middle position ($s = 0.5$) to further decrease the dynamics violation within this interval. With the approximation of the control and contact force to be linear function, their interpolated value at the mid-point is the average of the edge values $\bm{u}_{mid} = \frac{\bm{u}_i + \bm{u_}{i+1}}{2}$, $\bm{\lambda}_{mid} = \frac{\bm{\lambda}_i + \bm{\lambda}_{i+1}}{2}$. Together with the interpolated robot position $q_{mid}$, velocity $\dot{q}_{mid}$ and acceleration $\ddot{q}_{mid}$, a dynamics constraint is imposed at this collocation point
\begin{equation}
\begin{split}
        &D(q_{mid})\ddot{q}_{mid} + C(q_{mid},\dot{q}_{mid}) + G(q_{mid}) =\\ 
        &J(q_{mid})^{T}\bm{\lambda}_{mid} + B\bm{u}_{mid}
\end{split}
\end{equation}
By matching the cubic spline to the real trajectory at both knots and collocation, the dynamics constraint violation along this spline is significantly reduced. To get rid of the difficulty in formulating the undifferentiable self-collision avoidance constraint in 3 dimension environment, we assume at this stage the robot locomotion is in its sagittal plane. In addition, we constrain the relative distances of robot's internal joints to be always strictly away from the environmental features such that contact can only be made at robot's hands and feet.

   \begin{figure*}[!btp]
    \centering
    \includegraphics[scale = 0.6]{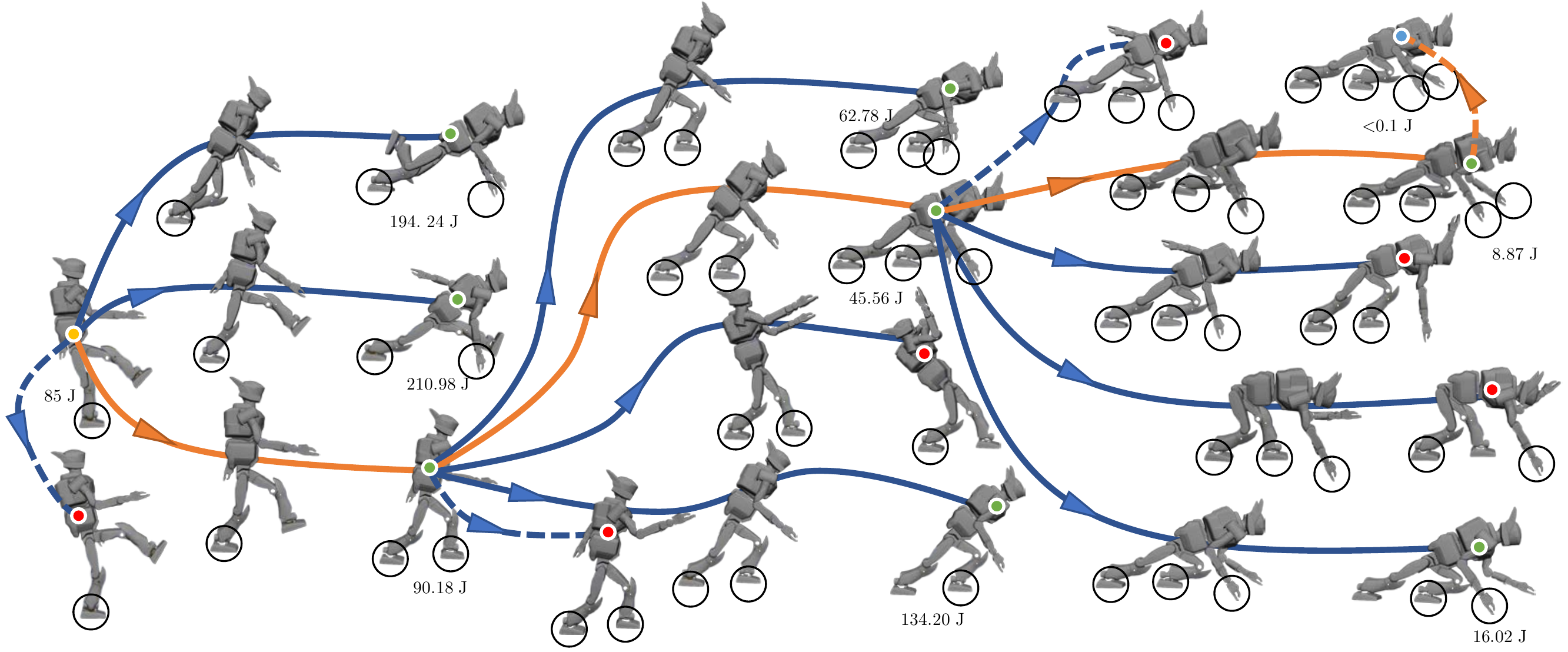}
    \caption{Contact transition tree on flat ground, starting from  initial kinetic energy \SI{85}{\J}. Solution takes a protective step, makes ground contact with both hands and uses inertia shaping to achieve a terminal stabilization. }
    \label{fig:Flat_KE85}
    \end{figure*}

The overall NLP that is solved is
\begin{equation}
\begin{aligned}
& \underset{h, \bm{x}_1, ..., \bm{x}_{N_d}, \bm{u}_1, ...,\bm{u}_{N_d}, \bm{\lambda}_1, ..., \bm{\lambda}_{N_d}}{\text{minimize}}
& & E_k(\bm{x}_{N_{d}}) \\
&  \hspace{12mm}\text{subject to}
& & (1) - (8)
\end{aligned}
\end{equation}
which we solve using the SNOPT library \cite{SNOPT}.

\subsection{Optimal Seed Initialization}
An initial guess is needed for the NLP solver to find a feasible and high-quality solution. This is a nontrivial challenge.  Our algorithm actually uses multiple initial guesses of increasing duration. For each duration, a smooth parabolic spline that obeys the initial and terminal constraints is generated.  We try solving the NLP when seeded from each of these initial guesses, and terminate when the first feasible solution is found.

For a given duration guess $T_i$, the initial guess satisfies the kinematic transition constraint and minimizes the violation of the dynamics constraint at each knot point. This procedure is as follows: 
\begin{enumerate}
    \item Take $\bm{x}_0 = [q_0^{T}, \dot{q}_0^{T}]^{T}$ as a starting point and compute a reference goal configuration $q_{ref}$ by assuming that $\dot{q}_0$ linearly decreases to zero at $T_i$ so $q_{ref} = q_0 + \dot{q}_0 T_i + \frac{1}{2}\ddot{q}_{ref} T_i^{2}$ where $\ddot{q}_{ref} = -\frac{\dot{q}_0}{T_i}$.
    \item Project $q_{ref}$ back to the constraint manifold to get the goal configuration $q_{g}$ and construct a parabolic curve with duration $T_i$ starting at $q_0$ and ending at $q_{g}$, with initial velocity $\dot{q}_0$.
    \item Discretize the parabolic spline into $n$ segments and interpolate the configuration, velocity and acceleration at edges (knots) of the segments.
    \item For each knot point, compute the left hand side of dynamics equation (\ref{eqn:dynamics}).  Solve for $\bm{u}$ and $\lambda$ in least squares fashion by multiplying the l.h.s. by the pseudo-inverse of $\left[B,  \hspace{0.75mm} J_\sigma(q)^{T}\right]$ where $J_\sigma$ is the Jacobian matrix of active contacts in the post-impact contact mode $\bm{\sigma}$.
\end{enumerate} 

The trajectory duration $T_i$ directly affects the acceleration of the parabolic curve.   Smaller values of $T_i$ generally yield larger accelerations, and hence more extreme control and contact forces.  We uniformly explore a  range of durations $T$ bounded between $\left[T_{min}, T_{max}\right]$, divided uniformly into $N_{tot}$ points.  Our optimizer explores these options via brute force under increasing duration $T_i$ until a feasible solution has been found or all options are exhausted. 

\subsection{Impact mapping}
\label{sec:ImpactMapping}
Impact happens when a contact is added, and we assume that it is inelastic and instantaneously changes the pre-impact velocities $\dot{q}^-$ to post-impact velocities $\dot{q}^+$ through an infinitesimal time duration.  $\dot{q}^+$ must satisfy the goal contact mode holonomic constraint, so we can calculate the impulse $\bar{\lambda}$ and post-impact state by solving the linear equation \cite{Hurmuzlu1994}. \begin{equation}
    \left[
    \begin{array}{cc}
    D(q) & -J_\sigma(q)^{T}\\
    J_\sigma(q) & \bm{0}\\
    \end{array}
    \right]
    \left[\begin{array}{c}
    \dot{q}^{+}\\
    \bar{\lambda}
    \end{array}
    \right]=\left[\begin{array}{c}
    D(q)\dot{q}^{-}\\
    \bm{0}.
    \end{array}
    \right]
    \end{equation}

\begin{table}[tbp]
\caption{Parameters used in experiments}
\centering
\begin{tabular}{ ccccll } 
\toprule 
\multicolumn{2}{c}{System Parameters} & \multicolumn{2}{c}{Optimization Coefficients } & \multicolumn{2}{c}{Tolerances}\\
\midrule
$n$ & 13  &$T_{min}$ & 0.25\,s&  $\epsilon_{E_k}$&\SI{0.1}{\J}\\            
$m$ &10 & $T_{max}$&3.5\,s         &  $\epsilon_{ct}$&0.025\, $m$\\            
$l$& 12 &$N_{tot}$&40&   & \\                
$\mu$ & 0.35   &$N_d$&8&    &  \\             
\bottomrule
\end{tabular}
\label{tab:coeff}
\end{table}

\begin{figure}[tbp]
    \centering
    \includegraphics[scale = 0.45]{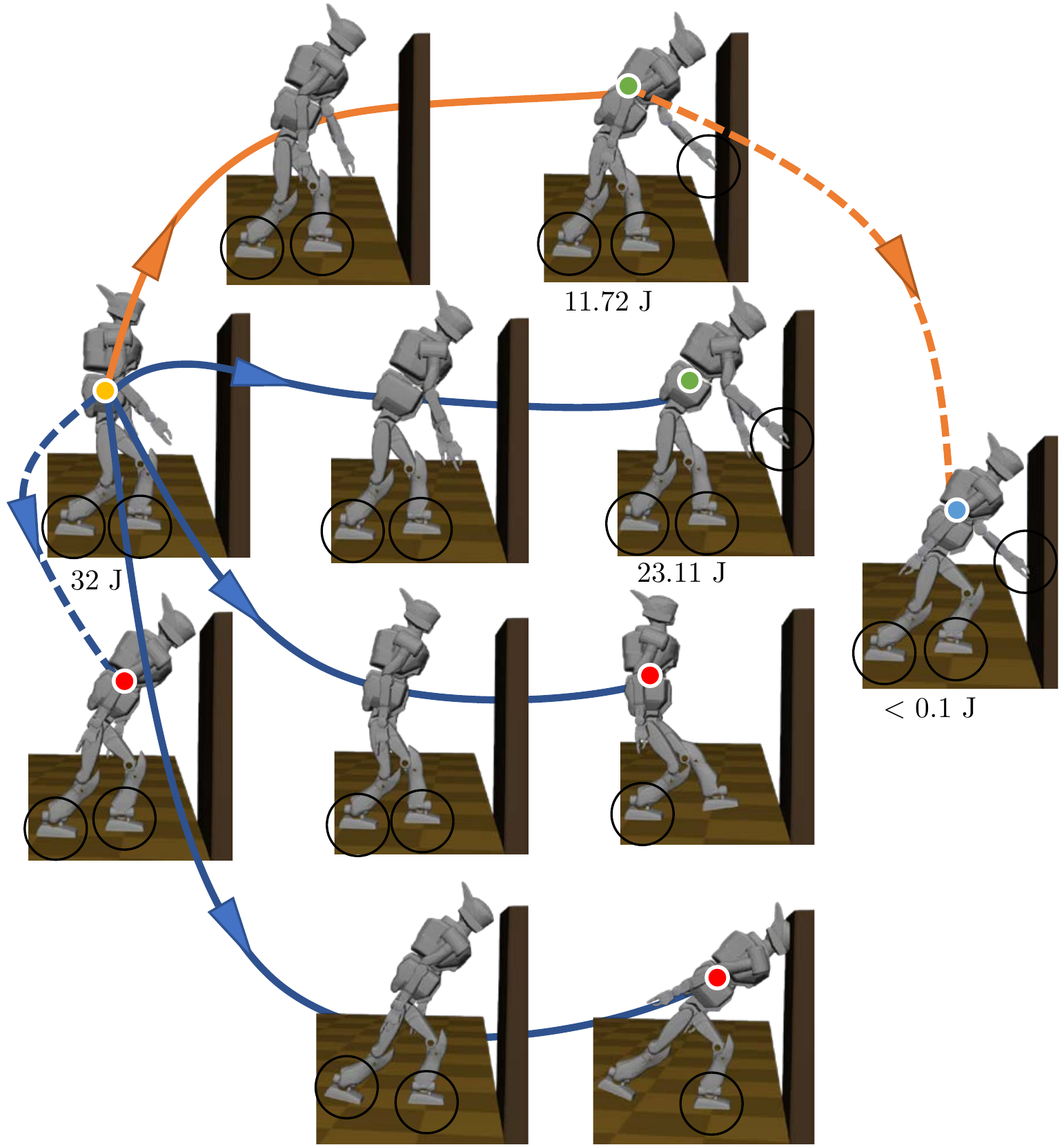}
    \caption{Contact transition tree with vertical wall, yielding a hand contact strategy.}
    \label{fig:Vert_Highlight}
    \end{figure}
    
\section{Experimental Evaluation}
We evaluate the effectiveness of the proposed method with a simulated planar model of the HRP-2 robot under varying initial disturbed robot states and environment shapes. All experiments are conducted on a 64-bit Intel Quad-Core i7 2.50GHz workstation with 8GB RAM. The computational time in solving the NLP takes around $3\sim 5$ min given a suitable initial seed. Robot parameters and optimization coefficients used in our experiment are listed in Tab.~\ref{tab:coeff}.

\subsection{Multi-Contact Fall Mitigation}

This subsection demonstrates the capability of the proposed algorithm to generate complex multi-contact stabilization strategies in different environments.

Fig.~\ref{fig:Flat_KE85} shows the contact transition tree produced by our algorithm with a large \SI{85}{\J} disturbance on flat ground.  There is no stabilizing self-motion trajectory with either 0, 1, or 2 contact switches, so the tree is expanded until it finds a solution at depth~3. The optimal contact sequence for this case is: protective stepping + two-hand contact + inertial shaping, specifically RF$\to$RF/LF$\to$RF/LF/LH$\to$RF/LF/LH/RH. 

Fig.~\ref{fig:Vert_Highlight} shows the contact transition tree produced for the robot pushed toward a vertical wall with an initial two-foot contact mode.  There are no self-motion trajectories at depth~0, but it finds a solution using hand contact, yielding a contact sequence LF/RF$\to$LF/RF/LH. 


\subsection{Change of Strategy with Increasing Initial Energy}
This subsection demonstrates how our algorithm can explore the change in fall mitigation strategy necessary to handle pushes of increasing severity. We choose 8 initial states whose kinetic energies increase in an evenly spaced pattern all the way up to the extreme case.

\begin{figure}[tbp]
    \centering
    \includegraphics[scale = 0.4]{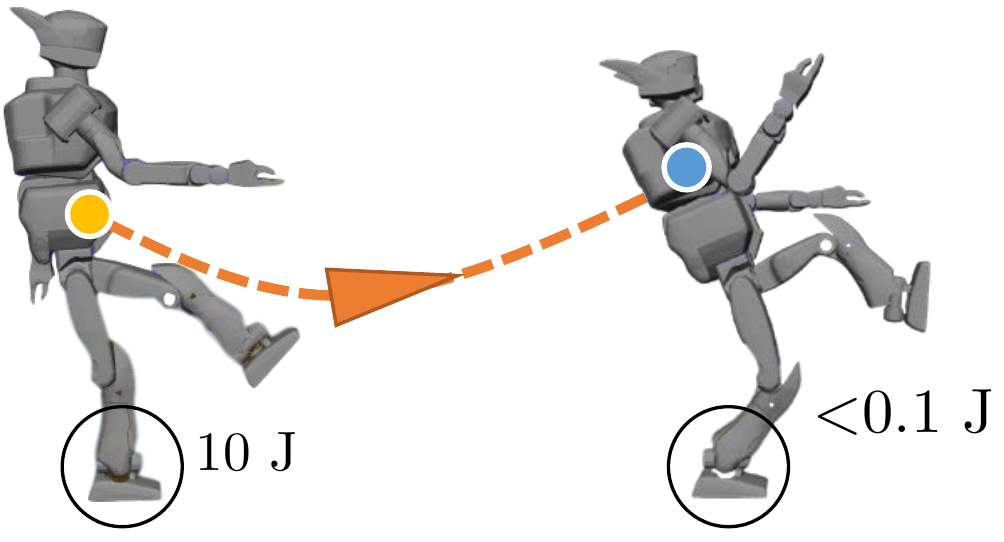}
    \caption{Inertial shaping with initial kinetic energy \SI{10}{\J}.}
    \label{fig:Flat_KE10}
    \end{figure}

\begin{figure}[tbp]
    \centering
    \includegraphics[scale = 0.5]{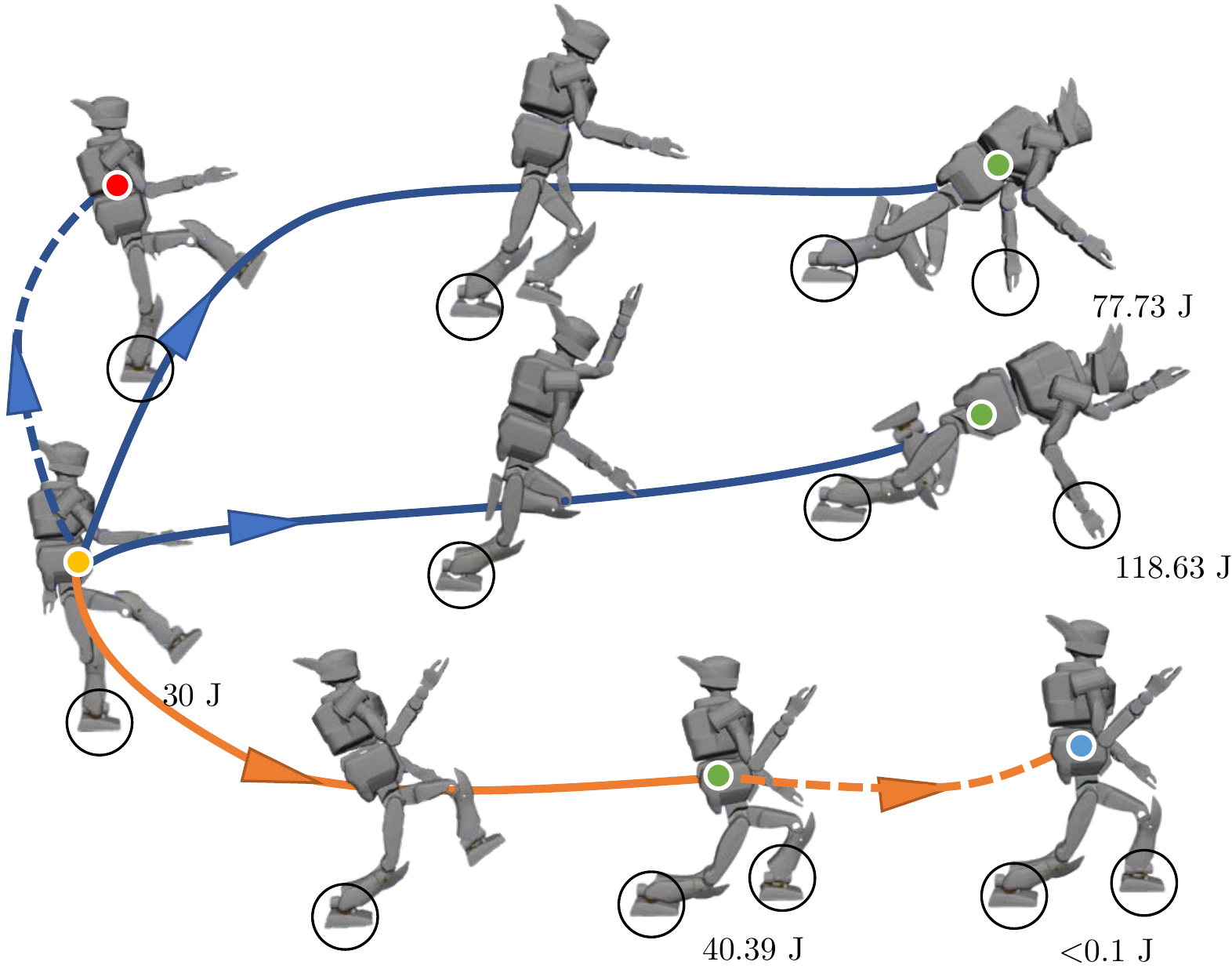}
    \caption{Protective stepping with initial kinetic energy \SI{30}{\J}.}
    \label{fig:Flat_KE30}
    \end{figure}
    
\begin{figure}[tbp]
    \centering
    \includegraphics[scale = 0.4]{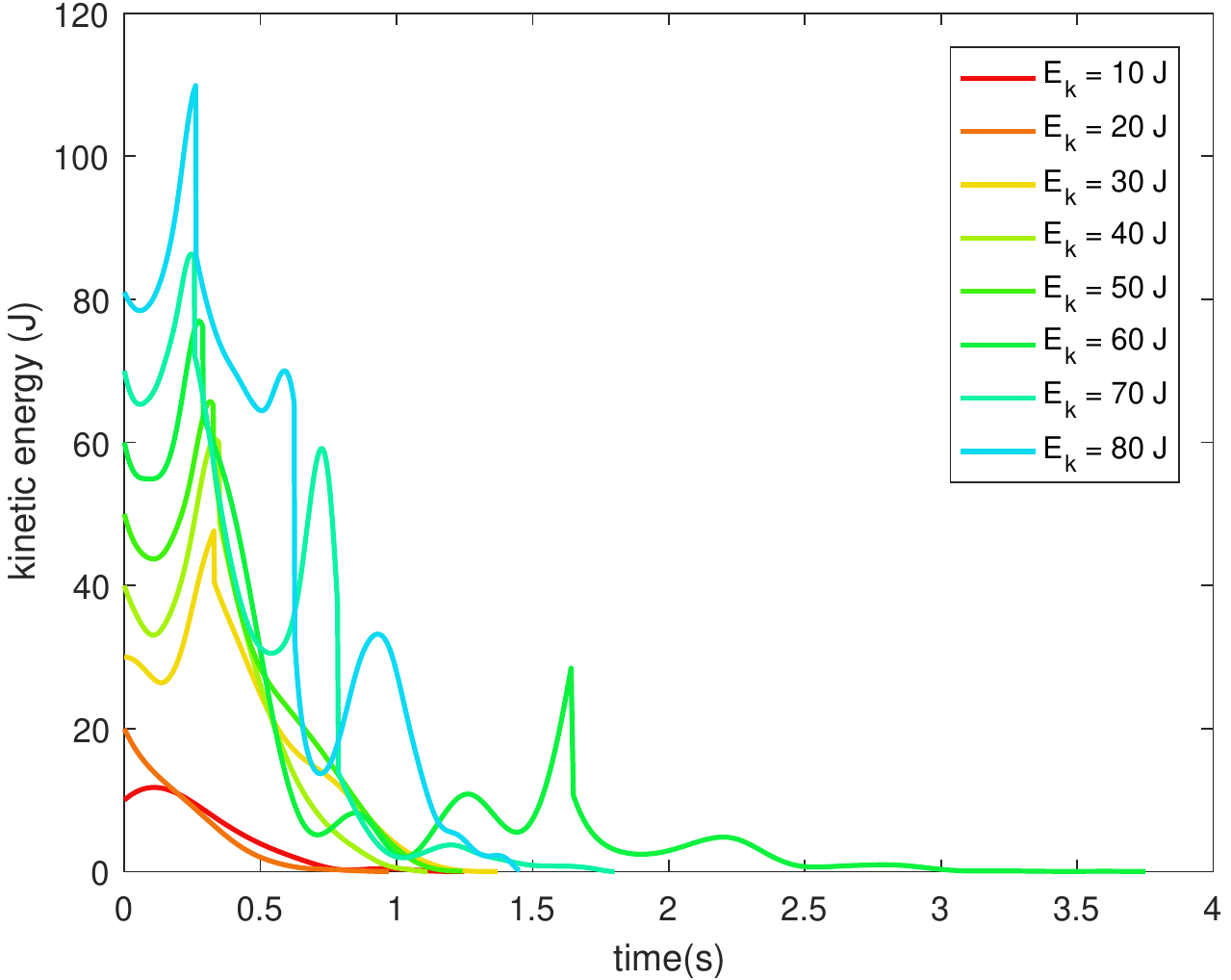}
    \caption{Kinetic energy trajectories with initial KE varying from \SI{10}{\J} to \SI{80}{\J}}
    \label{fig:Kinetic_Energy}
    \end{figure}

\begin{enumerate}
    \item $E_k(\bm{x}_{\bm{0}}) =$ \SI{10}{\J}, \SI{20}{\J}:
    The robot is able to dampen its momentum with inertial shaping, so its contact transition tree has only a single node (Fig.~\ref{fig:Flat_KE10}).

    \item $E_k(\bm{x}_{\bm{0}}) =$ \SI{30}{\J}, \SI{40}{\J}, \SI{50}{\J}: 
    Inertial shaping cannot stabilize the robot, and protective stepping is needed  (Fig.~\ref{fig:Flat_KE30}).
    
    \item $E_k(\bm{x}_{\bm{0}})=$ \SI{60}{\J}, \SI{70}{\J}, \SI{80}{\J}: 
    Neither inertial shaping nor protective stepping are sufficient. For these cases, our algorithm explores until depth 2, where hand contact enables successful stabilization.  The contact transition trees for these cases are similar to Fig.~\ref{fig:Flat_KE55}.
\end{enumerate}
Fig.~\ref{fig:Kinetic_Energy} lists the kinetic energy trajectories for all eight cases. All KE at ending time have been reduced to zero.

In addition, we further test the proposed algorithm in one extreme case ($E_k(\bm{x}_{\bm{0}})>$\SI{150}{\J}). When the initial  $E_k(\bm{x}_{\bm{0}})$ is extremely large, the planning of optimal contact sequence will fail to plan the stabilization strategies. This failure is due to constraints on the joint torque bounds and contact force feasibility. When the joint torque is not larger enough to maintain the contact holonomic constraints on position and velocity, the supportive normal contact force will need to be negative to drag the contact point on the contact surface. This negative contact force violates its feasibility constraints, thus preventing the optimal solution being computed. 

\section{CONCLUSION}
Our multi-contact planner for humanoid fall mitigation unifies inertial shaping, protective stepping, and hand contact strategies.  The planner optimizes both the contact sequence and the robot state trajectories using a contact transition tree search.  A greedy minimization of kinetic energy tends to find solutions with few contact changes and very little backtracking. An efficient method to generate initial seeds for trajectory optimization facilitates convergence. 
Experiments demonstrate show that our proposed algorithm can generate complex stabilization strategies for a simulated humanoid under varying initial pushes and environment shapes.

Despite these promising results, the algorithm is currently not suitable for  real-time use. Each self or transition optimization takes approximate $3\sim 5$ min provided promising initial seeds.  However, when unsatisfactory seeds are generated, the optimization solver will suffer from numerical difficulties and considerably large computation time will be taken. To address this, in the future we hope to use our optimizer generate large databases of optimal trajectories, and then train a machine learning model to rapidly predict the fall mitigation strategy.  Moreover, the current robot model is implemented in the HRP-2 robot's 2D sagittal plane, and we plan in the near future to implement a 3D version.  Finally, although greedy energy minimization works well for relatively simple environments, more complex environments may require increases of kinetic energy in order to stabilize the robot (for example, the robot may need to jump over a large gap).

\addtolength{\textheight}{-0.5cm}   

\bibliographystyle{IEEEtran}
\bibliography{references}

\end{document}